\documentclass[conference]{IEEEtran}

\usepackage{amsmath,graphicx,algorithm,algpseudocode,amsfonts}
\usepackage{epstopdf}
\graphicspath{{Figures/}}
\usepackage[flushleft]{threeparttable}

\algnewcommand\algorithmicinput{\textbf{Input:}}
\algnewcommand\INPUT{\item[\algorithmicinput]}
\algnewcommand\algorithmicoutput{\textbf{Output:}}
\algnewcommand\OUTPUT{\item[\algorithmicoutput]}
\algnewcommand{\LineComment}[1]{\State \(\triangleright\) #1}

\begin{document}

\title{An Effective Feature Selection Method Based on Pair-Wise Feature Proximity for High Dimensional Low Sample Size Data}

\author{\IEEEauthorblockN{S L Happy}
\IEEEauthorblockA{Department of Electrical Engineering, \\Indian Institute of Technology \\Kharagpur, India\\Email: happy@iitkgp.ac.in}
\and
\IEEEauthorblockN{Ramanarayan Mohanty}
\IEEEauthorblockA{Advanced Technology \\Development Center, \\Indian Institute of Technology \\Kharagpur, India\\Email: ramanarayanmohanty@iitkgp.ac.in}
\and
\IEEEauthorblockN{Aurobinda Routray}
\IEEEauthorblockA{Department of Electrical Engineering, \\Indian Institute of Technology \\Kharagpur, India\\Email: aroutray@iitkgp.ac.in}}
\maketitle

\begin{abstract}
Feature selection has been studied widely in the literature. However, the efficacy of the selection criteria for low sample size applications is neglected in most cases. Most of the existing feature selection criteria are based on the sample similarity. However, the distance measures become insignificant for high dimensional low sample size (HDLSS) data. Moreover, the variance of a feature with a few samples is pointless unless it represents the data distribution efficiently. Instead of looking at the samples in groups, we evaluate their efficiency based on pair-wise fashion.
In our investigation, we noticed that considering a pair of samples at a time and selecting the features that bring them closer or put them far away is a better choice for feature selection.
Experimental results on benchmark data sets demonstrate the effectiveness of the proposed method with low sample size, which outperforms many other state-of-the-art feature selection methods.
\end{abstract}

\IEEEpeerreviewmaketitle

\section{Introduction}	\label{sec:intro}

In this age of information, high dimension data with low sample size are very common in various areas of science \cite{HDLSS_DSP2015}. 
In supervised classification problems, the classification performance is mostly determined by the inherent class information possessed by the features. Hence, it is logical to include more number of features to improve the discriminating ability.
In this way, most of the practical machine learning tasks deal with high dimensional data while the number of labeled data are far less than its dimensionality. The feature space for such data is almost empty, and the curse of dimensionality causes the distance measure to become uniform \cite{tang2014feature}.  In addition, the sparsity of labeled instances in high dimensional feature space adversely affects the classification performance as well. 

Feature selection is a process of selecting an optimal subset of features from the input feature set based on a selection criterion \cite{yu2004efficient}. Thus, it reduces the data dimensionality by removing redundant features and improves the time and space complexity of the data. In addition, it reduces the risk of over-fitting which is very common in high dimensional data analysis. These algorithms can be categorized as supervised, unsupervised or semi-supervised \cite{zhao2013similarity} based on their utilization of label information. Different criteria functions have been proposed in the literature to evaluate the goodness of features, such as mutual information (MutInf) \cite{mutualfs2002}, Fisher score (FS) \cite{duda2001pattern},  feature selection via concave minimization (FSCM) \cite{bradley1998feature}, ReliefF \cite{liu2007computational}, Laplacian score (LS) \cite{laplacianScoreNIPS2005}, trace ratio criterion (TRC) \cite{nie2008trace}, spectral feature selection (SPEC) \cite{specICML2007}, infinite feature selection (IFS) \cite{ifsICCV2015} etc. They have demonstrated excellent performance in real-world applications.

The key to obtaining the suitable subset of features depends upon the selection criteria. Algorithms, such as FS and ReliefF, optimize the sample separability, whereas LS preserves sample similarity in the local neighborhood \cite{laplacianScoreNIPS2005}. MutInf considers the mutual information between the distribution of a feature as the selection criterion \cite{mutualfs2002}. SPEC selects the features by analyzing the spectrum of the graph induced from the proximity matrix \cite{specICML2007}. 
However, all these methods evaluate each feature independently and select the top ones based on the utility of features. Such heuristic algorithms neglect the combined performance of multiple features which leads to the selection of a suboptimal subset of features \cite{liu2017new}. Thus, it is entirely possible that the performance with two best scoring features may be lower than the performance of any other two features combined. However, finding the global optimal solution is an NP hard problem and very challenging.


Feature selection methods like generalized Fisher score (GFS) \cite{GFS_2011}, TRC and IFS try to globally optimize feature subset to maximize the subset level score. GFS jointly selects features, which maximize the lower bound of traditional Fisher score. Similarly, IFS considers each feature as a node in the affinity graph and assigns a score to each by taking into account all the possible feature subsets as paths on a graph \cite{ifsICCV2015}. 
These methods eliminate the redundant features while considering the combination of features, which gives an advantage over independent feature evaluation methods.

The graph based methods need a suitable number of instances to learn the graph structure, failing which results in inferior performance. Higher accuracy demands sufficient labeled training data, however, annotating data for real-world applications is an expensive and time consuming job \cite{luo2013vector}. The methods that compute the features independently are less affected by the low sample size. Nevertheless, the performance of most of the selection criteria decreases with a low number of sample instances. 

We address the issues that arise when there is a lack of labeled data samples. When the available samples are less, the only way to select the best features is to assign scores based on how close they are to the samples of the same class while keeping maximum distance from other class samples. We propose a naive way of selecting features which involves the combinational feature selection followed by the heuristic approach of score assignment to each feature. It takes the advantages of selecting a group of features based on the pair-wise proximity in feature values and the lower computational complexity of heuristic search for assigning scores and ranking the features. 
Instead of using the whole feature vector for distance measurement, we use a subset of the original feature set. Thus, features, responsible for bringing the points of the same class closer while keeping a safe distance from the other class instances, can be found based on the distance between each pair of training instances. Here the basic assumption is that the optimal feature set minimizes the within-class distance, while maximizing the between-class distances for each pair of samples. The proposed pair-wise feature proximity (PWFP) based feature selection method is compared with other literature and evaluated extensively.

\subsection{Notation}
In a supervised feature selection scenario, the algorithm is provided with the data and the corresponding class label. Suppose the data set consists of $n$ number of $d$-dimensional points $(x_i=[x_i^1,x_i^2,...,x_i^d] \in \mathbb{R}^d)$, given by $\{(x_i,y_i)\}_{i=1}^n$. Here $y_i \in \{1,2,...,c\}$ represents the class label of the corresponding data. The problem of feature selection aims at finding the feature subset which carries the maximum information to classify the features into accurate classes. 
Further, we denote the total data matrix as $X=[x_1,x_2,...,x_n]\in \mathbb{R}^{d\times n}$, and $f^i = [f^{i1},f^{i2},...,f^{in}]$ represents the $i$th row of the matrix $X$. Without loss of generality, we assume that X has been centered with zero mean, i.e., $\sum_{i=1}^n{x_i} = 0$.

\section{Proposed Method}		\label{sec:proposed_method}
This section describes the proposed PWFP feature selection method in detail. Before going to the details of the proposed method, a brief discussion of the existing methods is provided which stand as the ground for our arguments.

The feature selection problem may be formulated as finding $m$ features out of $d$ dimensions, which will provide the optimum classification accuracy. Thus, it involves $\left(\begin{smallmatrix} d \\ m\end{smallmatrix}\right)$ candidates and becomes a combinational optimization problem, the solution to which is very challenging. Usually, heuristic strategy \cite{duda2001pattern} is used to find the best features by evaluating each feature independently. In this case, the evaluation is carried for each $d$ features (thus, $d$ candidates), and the top $m$ features are selected. Most of the available feature selection methods in the literature try to impose different evaluation criteria to obtain suitable performance.

Two widely used filter-based feature selection methods are FS and LS. The evaluation criterion used in FS \cite{duda2001pattern} maximizes the between-class variance while minimizing the within-class variance. Thus, the FS is formulated as
\begin{equation}
F(f^i) = \frac{\sum_{k=1}^c n_k (\mu_k^i - \mu^i)^2}{\sum_{k=1}^{c}n_k (\sigma_k^i)^2}
\end{equation}
where $\mu_i$ is the overall mean of $i$th feature, $\mu_k^i$ is the $i$th feature mean of $k$th class, $\sigma_k^i$ is the $i$th feature variance of $k$th class, and $n_k$ is the number of samples of $k$th class. 
On the other hand, LS \cite{laplacianScoreNIPS2005} takes into account the similarity or the closeness of the data points for feature evaluation. It constructs a graph to reflect the local geometric structure and seeks the features which respect that graph.
It tries to find the feature that minimizes the difference of the data points from the same class or the closely situated points, while possessing a high global variance. Given the similarity score between points $x_j$ and $x_k$ as $S_{jk}$, the LS is calculated by,
\begin{equation}
L(f^i) = \frac{\sum_{j,k} (f^{ij}-f^{ik})^2 S_{jk}}{Var(f^i)}
\end{equation}
where $Var(.)$ represents the variance of the feature. Usually, Euclidean distance is used to find the similarity for graph construction. However, the computation of Euclidean distance has its inherent problems. When the dimension of the data points is very high, the distance metrics become meaningless \cite{high_dimen_distance_TAKDE_2009}. Thus, finding close points in $\mathbb{R}^d $ is a difficult task and it affects the similarity measures based on the Euclidean distance and the local graph structure as well. The Euclidean distance is given by, 
\begin{equation}
dist^2(x_j,x_k) = \sum_{i=1}^d (x_j^i - x_k^i)^2  = (x_j - x_k)^\mathrm{T} (x_j - x_k)
\end{equation}
which uses the square of the difference of each feature dimension to compute the distance. Consider two points closely situated in high dimensional space. The presence of noise in any one dimension will increase the distance between these pair of points. Even normalization of features does not help much to alleviate the issue. And the computation of FS, LS, and other such methods are also affected. 

When the dimensionality of data points is very high compared to the number of samples available for training, these methods experience a few disadvantages.
First, the high dimensional data are almost empty and the computed variance carries no meaning unless the samples represent the data distribution properly. 
Second, the distance measure for these high dimensional data becomes almost uniform. Therefore, the similarity based methods are adversely affected.
Third, the graph-structures formed with an insufficient number of instances are inaccurate to represent the feature manifold.
Finally, the features selected by these heuristic algorithms are sub-optimal as each feature is computed independently and the effects of the combination of more than one feature are neglected. For example, two features may have low individual scores, however, their combined score may be very high. In this case, FS will not select either of them, although they should be selected for accurate classification. 

Our formulation is based on the idea of Fisher score computation. We propose a naive way of selecting features based on pair-wise feature similarity. 
Feature variance should be low for the points belonging to the same class, while it should be high for points belonging to different classes. Fisher criterion uses similar logic, while considering all the samples of different classes altogether. However, the computation of mean and variance are affected with low sample size. Therefore, we use the pair-wise feature similarity to select the appropriate features.

A feature is said to be a `good' feature if it keeps the samples of the same class close, while keeping the points from different classes far away. Alternatively, if we consider a pair of points, a good feature should have the following properties. 1) For the pair that belongs to the same class, the values of feature should be close. 2) For the pairs belonging to different classes, the feature should be able to differentiate the classes easily. Thus, we seek the feature dimensions along which the point pairs are very close for the same class and very far for different classes.

Lets define $p_{jk}=[b_1,b_2,...,b_d]^\mathrm{T}, b_i \in \{0,1\}$, with $b_i=1$ as the features along which a pair of points $(x_j,x_k); y_j = y_k$ are close to each other. Thus, the $b_i=1$ features in $p_{jk}$ are the features along which the pair-wise with-in variance is minimum. We can choose these features by sorting the distance between individual features in ascending order and selecting the first few features. One naive way of doing so is to choose the features satisfying the following optimization problem,
\begin{align} \label{findp}
\max_{p_{jk}} &\; p_{jk}^Tp_{jk} \nonumber \\
s.t. &\; |x_j-x_k|^\mathrm{T}p_{jk} < \tau
\end{align}
where $\tau$ is a threshold. Here we use manhattan distance metric as it has been reported to have significant performance \cite{aggarwal2001surprising} for high dimensional data.
Suppose, we need to keep $\beta$ number of features out of $d$, which are close for the pair  $(x_j,x_k)$. This even makes the selection process more easy.
\begin{align} \label{equ:find_p}
\min_{p_{jk}} \; |x_j-x_k|^\mathrm{T}p_{jk}\; ; \quad 
s.t. \quad p_{jk}^Tp_{jk} = \beta
\end{align}
The manhattan distance between $x_j$ and $x_k$ is $\sum_{i=1}^d |x_j^i - x_k^i| = |x_j - x_k|^\mathrm{T} \mathbf{1} $, where $\mathbf{1} $ is a vector of ones. Thus, we can interpret the term $|x_j-x_k|^\mathrm{T}p_{jk}$ as a distance measure with a sub-set of features for which $b=1$.

Similarly, let $q_{jk}=[b_1,b_2,...,b_d]^\mathrm{T}, b_i \in \{0,1\}$ be the features along which the pair $(x_j,x_k); y_j \ne y_k$ are farthest if $b_i=1$. A similar way of finding the features that discriminate the points from different classes can be given by,
\begin{align} \label{equ:find_q}
\max_{q_{jk}} \;|x_j-x_k|^\mathrm{T}q_{jk} \; ; \quad
s.t \quad  q_{jk}^Tq_{jk} = \beta
\end{align}

We collect the information from all such possible pairs which are represented by $P$ and $Q$ respectively, given by
\begin{align}\label{equ:calcPQ}
P = \frac{1}{N_p} \sum_{j,k; y_j=y_k} p_{jk} \nonumber \\
Q = \frac{1}{N_q} \sum_{j,k; y_j \ne y_k} q_{jk} 
\end{align}
where $N_p$ and $N_q$ are normalization factors. We use $N_p = \sum_{k=1}^c \left(\begin{smallmatrix} n_k \\ 2\end{smallmatrix}\right)$ and $N_q = \sum_{j,k;j \ne k} n_jn_k$. 
Here $P$ and $Q$ represents the normalized histogram of features based on their contribution toward closeness or discriminating power between different classes. 


Now, we seek the feature dimensions which are present in both $P$ and $Q$. To be specific, a good feature should be capable of discriminating between points from different classes while bringing the points of same class closer. Therefore, a good feature is a feature that has higher probability occurrence in both $P=[p^1,p^2,...,p^d]$ and $Q=[q^1,q^2,...,q^d]$. A reasonable criterion for choosing a good feature is to minimize the object function given by,

\begin{equation}\label{proposedScore}
\min_i \Big|\frac{p^i - q^i}{p^i + q^i}\Big|
\end{equation}

Apparently, once $P$ and $Q$ are found, the problem (\ref{proposedScore}) can be solved by element-wise operation of these two vectors. Furthermore, the term $S(i)=\Big|\frac{p^i - q^i}{p^i + q^i}\Big|$ can be used as a score for selecting features. The top $m$ features are those with the lowest scores. The algorithm for the proposed feature selection method is provided in algorithm \ref{alg:proposedAlg}.

\begin{algorithm}[!b]
	\caption{Algorithm for feature selection based on the pair-wise feature proximity (PWFP)}
	\label{alg:proposedAlg}
	\begin{algorithmic}[1]
		\INPUT Training data $\{(x_i,y_i)\}_{i=1}^n$, the selected feature number $m$, and the parameter $\beta$
		\OUTPUT The selected feature subset
		\For {$\forall (x_j,x_k) \in X$}
		\If {$y_j = y_k$}
		\State Compute $p_{jk}$ using (\ref{equ:find_p})
		\Else 
		\State Compute $q_{jk}$ using (\ref{equ:find_q})
		\EndIf
		\EndFor
		\State Compute $P$ and $Q$ using (\ref{equ:calcPQ}).
		\State Calculate the score of each feature based on $S(i)=\Big|\frac{p^i - q^i}{p^i + q^i}\Big|$
		\State Rank the features according to the scores in descending order
		\State Select the leading $m$ features
	\end{algorithmic}
\end{algorithm}

\begin{figure*}[t]
	\begin{minipage}[t]{1.0\linewidth}
		\centering
		\centerline{\includegraphics[width=7.25cm]{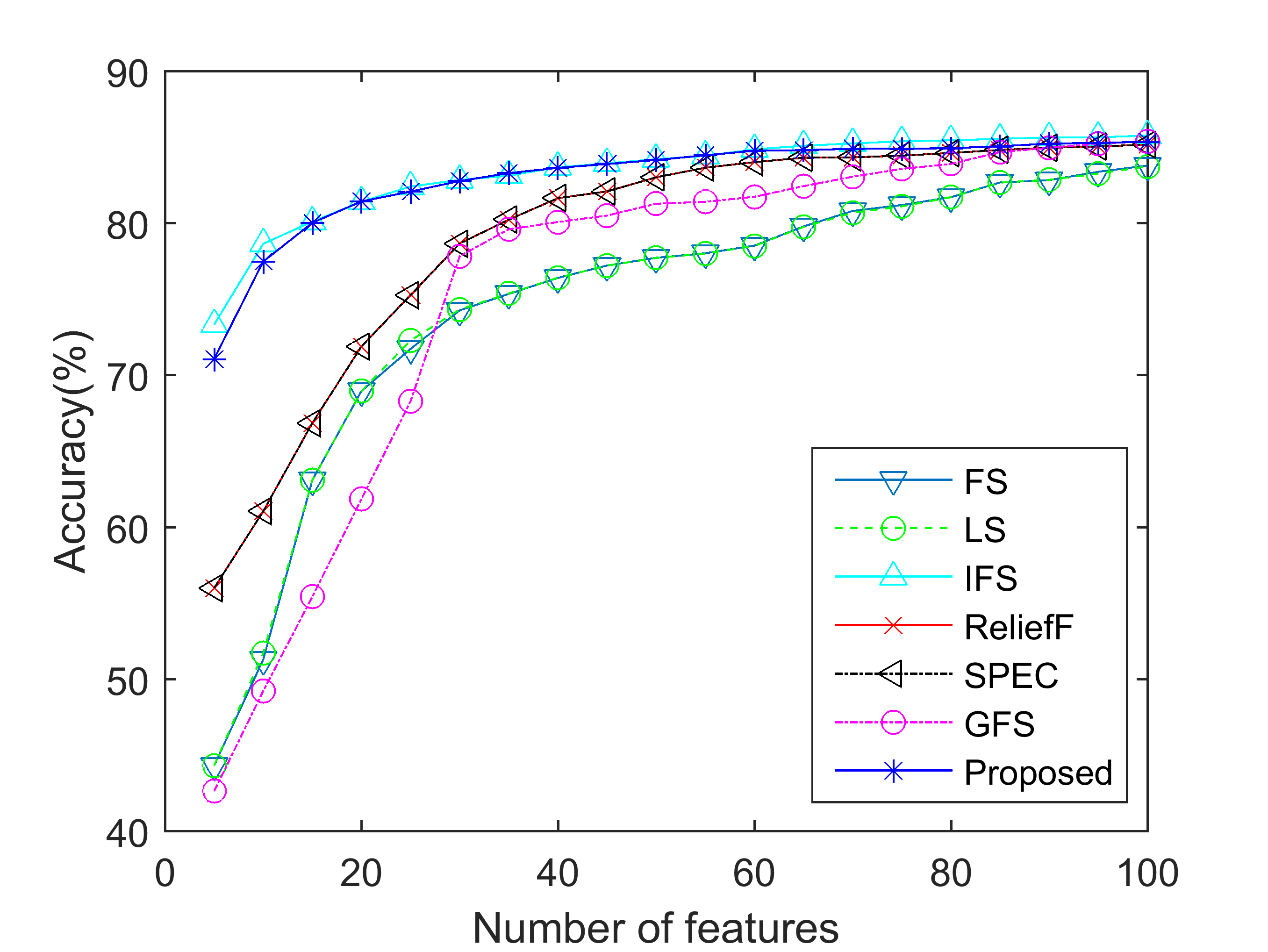},\includegraphics[width=7.25cm]{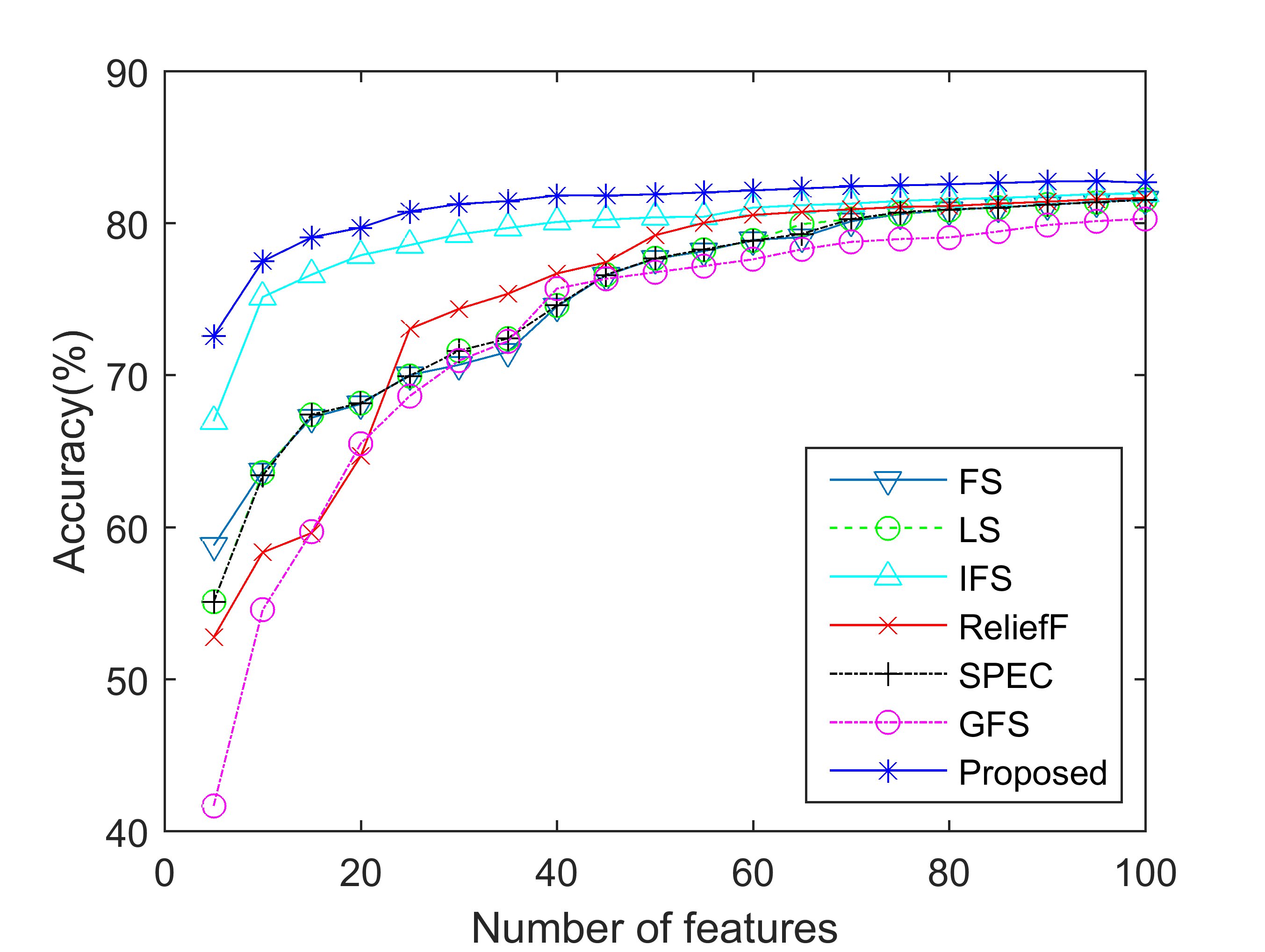}}
	\end{minipage}
	\caption{Comparison of performances of proposed (PWFP) and other methods in (a) Botswana (left) and (b) Salinas (right) database. [best viewed in color]}
	\label{fig:HSI_comparison}
\end{figure*}

As can be observed, we first select the optimal set of features for each pair of sample instances. These set of features from all such pairs are further brought together and the best features are selected based on the scores assigned to each feature. Thus, the proposed PWFP involves the combinational feature selection followed by a heuristic approach of score assignment to each feature. It takes the advantages of selecting a group of features based on the pair-wise proximity in feature values and the lower computational complexity of heuristic search for assigning scores and ranking the features. 

The first step in PWFP is similar to that of Fisher criterion. If we consider a pair of samples, then it is equivalent to selecting $m$ features using FS. 
However, FS considers all data together, while PWFP uses the pair-wise data. Unlike FS or LS, our method selects multiple features for each pair of data and combines them to select the best ones in a later stage. Therefore, the fear of suboptimal feature selection due to the evaluation of independent features is avoided. 
No graph representations are considered as we have assumed less number of available labeled samples. Furthermore, PWFP does not use similarity based on full feature set, which avoids over-fitting. 

\section{Experiments and Results}
\label{sec:result}

The performance of the PWFP is validated through the experiments conducted on several real-world data sets. 
A preprocessing step was carried out to normalize the samples to zero mean and unit variance. In all cases, a few samples were selected for training, while the rest were used for testing purpose. Linear support vector machines were used for classification purpose in all cases. We set the value of $\beta$ to be 10\% of the dimension in all the experiments.

\subsection{Hyperspectral data sets}
Hyperspectral images consist of hundreds of spectral bands to provide information about the properties of land cover. With a few annotations from the experts, segmentation and classification algorithms find the labels for the rest of the image. This is a special case of high dimensional low sample size data. 
In our experiments, we used Botswana\footnote{http://www.ehu.eus/ccwintco/index.php?title=\\Hyperspectral\_Remote\_Sensing\_Scenes} ($d=145,c=14$) and Salinas scene\footnotemark[\value{footnote}] ($d=204,c=16$). The training set was constructed with random selection of 10 sample points from each class and the rest were treated as test data. We report the average performance over ten iterations.

\begin{figure}[t]
	\begin{minipage}[b]{1.0\linewidth}
		\centering
		\centerline{\includegraphics[width=7.5cm]{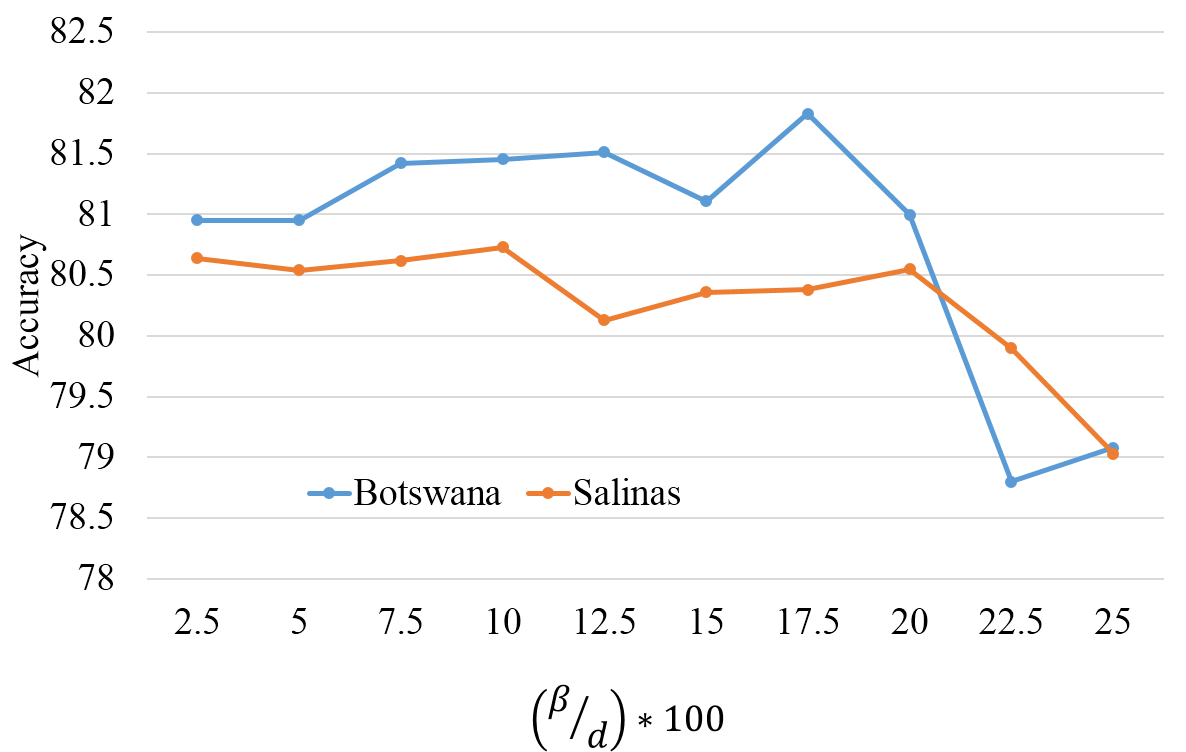}}
	\end{minipage}
	\caption{Effect of variation of $\beta$ on classification performance.}
	\label{fig:varyingBeta}
\end{figure}

The recognition accuracy with respect to the number of selected features of all the feature selection methods is provided in Fig. \ref{fig:HSI_comparison}. We observed that some of the existing methods performed almost equal on these data sets. The low sample size might be the reason behind different methods having similar properties. IFS performed the best among them. However, the PWFP achieved remarkable accuracy in Salinas database with a different number of selected features, while performing close to IFS in Botswana data set. 
The accuracy obtained with varying $\beta$ is illustrated in Fig. \ref{fig:varyingBeta}. As can be observed, the accuracy increases with increase of $\beta$, peaks, and then decreases. Empirically, we selected $\beta$ as 10\% of the feature dimension in all the experiments.

\subsection{Face Recognition}
\begin{figure}[t]
	\begin{minipage}[b]{1.0\linewidth}
		\centering
		\centerline{\includegraphics[width=7.5cm]{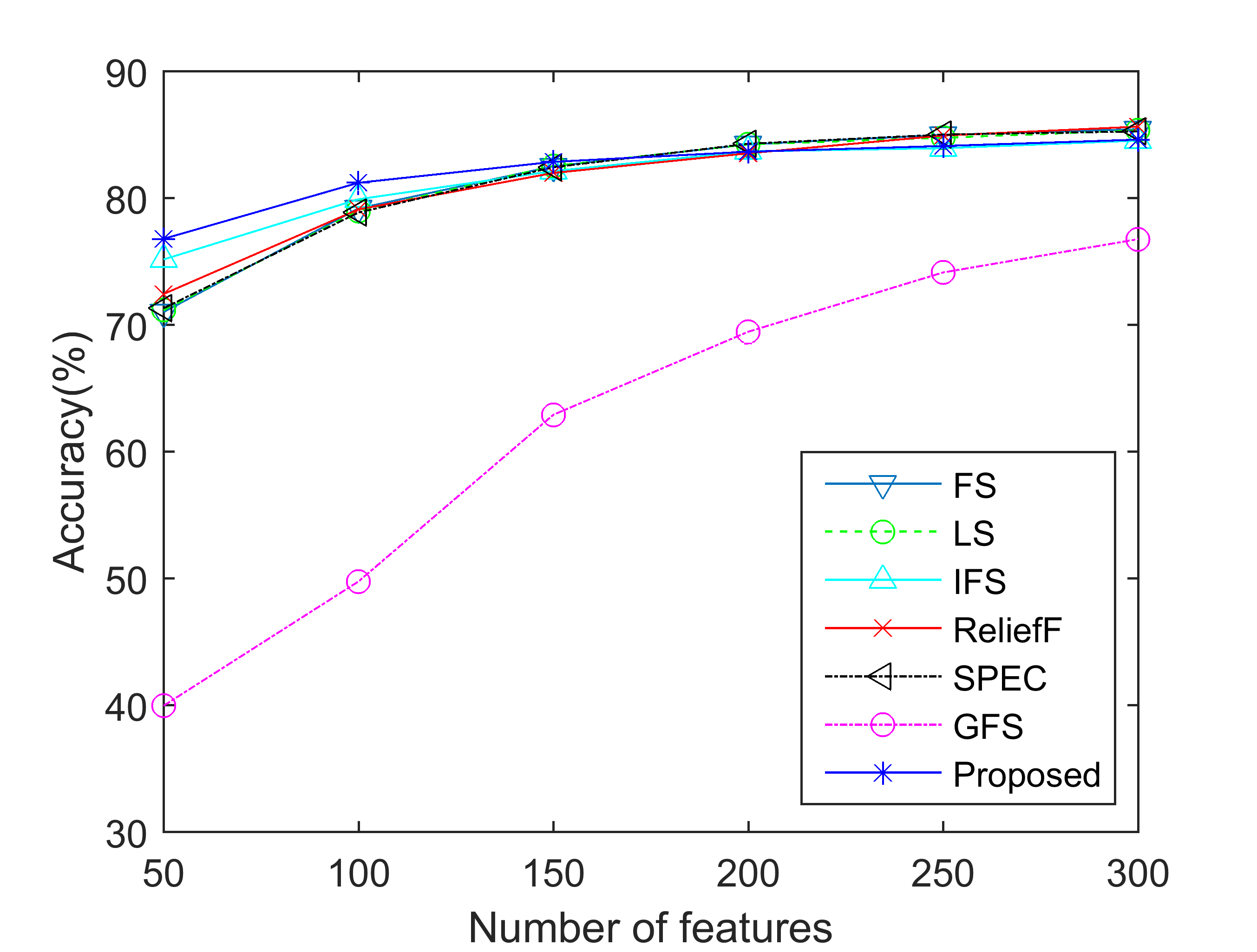}}
	\end{minipage}
	\caption{Performance of different feature selection methods on ORL data set.}
	\label{fig:orl}
\end{figure}

We carried out experiments on the ORL face recognition data set\footnote{http://www.cl.cam.ac.uk/research/dtg/attarchive/facedatabase.html}, which contains 10 images for each of the 40 participants ($c=40$). 
The face images were resized to $32\times 32$ and vectorized ($d=1024$). Randomly 5 images from each person were selected for training and the rest for testing. The average of 10 experiments with a different number of feature selection is provided in Fig. \ref{fig:orl}.

As can be seen, the PWFP outperforms the other feature selection methods when a few features are selected. 
GFS underperformed in this data set, while the other methods achieved almost equivalent accuracy. Lack of sufficient data for each class might be the reason for the failure of GFS algorithm.

\subsection{Other data sets}
A few publicly available databases having less number of sample instances with high dimension were used in our experiments.
The chosen data sets are Colon cancer diagnosis dataset\footnote{http://www.stats.uwo.ca/faculty/aim/2015/9850/microarrays/FitMArray\\/chm/Alon.html}, Lung cancer\footnote{http://www.pnas.org/content/98/24/13790/suppl/DC1}, protein\footnote{http://archive.ics.uci.edu/ml/datasets.html} , ionosphere\footnotemark[\value{footnote}], arcene\footnotemark[\value{footnote}], and tox-171\footnote{http://featureselection.asu.edu/old/datasets.php}.
For each data set, 10\% samples were randomly selected as training data and the rest were treated as test data. We repeated this procedure five times and the average performance is reported in Table \ref{table1}.

As can be seen in Table \ref{table1}, the data sets are arranged in increasing order of data dimensionality and the performance of the proposed method is compared with a few other methods that optimize the features globally. The accuracy achieved by PWFP is less when the data dimension is low (for protein and ionosphere). However, its performance goes higher as the dimensionality increases. PWFP achieved the best classification accuracy for the last four data sets. This validates the efficiency of the PWFP for high dimensional data with low sample size. 

\begin{table}[t]
	\centering
	\caption{Classification results on different data sets when 10\% data are used for training and the number of selected features is set to be 50\% of the dimensionality of the data. }
	\label{table1}
\fontsize{8}{8}\selectfont
	\renewcommand{\arraystretch}{1.5}
	\tabcolsep=2.5pt
	\begin{threeparttable}
	\begin{tabular}{c c c c c c c}
		\hline
		\textbf{Methods} & \textbf{protein} & \textbf{ionosphere} & \textbf{colon}  & \textbf{lung} & \textbf{TOX-171} & \textbf{arcene\tnote{*}} \\ 
		\begin{tabular}[c]{@{}c@{}}(samples,\\ dimensions)\end{tabular} & (116,20) & (351,34) & (62,2000) & \begin{tabular}[c]{@{}c@{}}(203,\\ 3312)\end{tabular} & \begin{tabular}[c]{@{}c@{}}(171,\\ 5748)\end{tabular} & \begin{tabular}[c]{@{}c@{}}(200,\\ 10000)\end{tabular} \\
		\hline
		IFS & 36.34 & 79.74 & 58.9 & 83.4 & 55.94 & 75.8 \\
		SPEC & \textbf{45.57} & 81.71 & 58.54 & 84.17 & 56.33 & 56 \\
		GFS & 43.65 & \textbf{81.96} & 60.72 & 75.05 & 57.12 & 79.6 \\
		Proposed & 43.26 & 78.96 & \textbf{64.36} & \textbf{84.72} & \textbf{57.77}& \textbf{81 }	\\
		\hline
	\end{tabular}
	\begin{tablenotes}
		\item[*] For arcene data set, 50\% data were used for training.
	\end{tablenotes}
\end{threeparttable}
\end{table}

\section{Conclusion} \label{conclusion}
In this paper, we propose a feature selection method based on pair-wise feature proximity. We use the closeness (or remoteness) of a feature dimension among a pair of points from the same (or different) class to select the efficient features for the class discrimination. 
The proposed method is analyzed extensively with a few high dimensional low sample size databases. It is found that the proposed method outperforms the existing algorithms when the database has a few samples with very high dimension.


\bibliographystyle{IEEEtran}
\bibliography{featureSelection}

\end{document}